\documentclass[sigconf]{acmart}
\usepackage{booktabs} %

\copyrightyear{2017} 
\acmYear{2017} 
\setcopyright{acmlicensed}
\acmConference{SCA '17}{July 28-30, 2017}{Los Angeles, CA, USA}\acmPrice{15.00}\acmDOI{10.1145/3099564.3099581}
\acmISBN{978-1-4503-5091-4/17/07}

\usepackage{rotating}
\usepackage{color}
\definecolor{finalcolor}{rgb}{0.0,0.5,0.0}
\newcommand{\FINAL}[1]{#1}

\title[Production-Level Facial Performance Capture Using Deep CNNs]{Production-Level Facial Performance Capture\\Using Deep Convolutional Neural Networks}

\author{Samuli Laine}
\affiliation{\institution{NVIDIA}}
\author{Tero Karras}
\affiliation{\institution{NVIDIA}}
\author{Timo Aila}
\affiliation{\institution{NVIDIA}}
\author{Antti Herva}
\affiliation{\institution{Remedy Entertainment}}
\author{Shunsuke Saito}
\affiliation{Pinscreen}
\affiliation{University of Southern California}
\author{Ronald Yu}
\affiliation{Pinscreen}
\affiliation{University of Southern California}
\author{Hao Li}
\affiliation{\mbox{USC Institute for Creative Technologies}}
\affiliation{University of Southern California}
\affiliation{Pinscreen}
\author{Jaakko Lehtinen}
\affiliation{\institution{NVIDIA}}
\affiliation{\institution{Aalto University}}
\renewcommand{\shortauthors}{Laine et al.}

\begin{document}
\newcommand{\figdiagram}{
\begin{figure}[t]
\centering
\includegraphics[width=\linewidth]{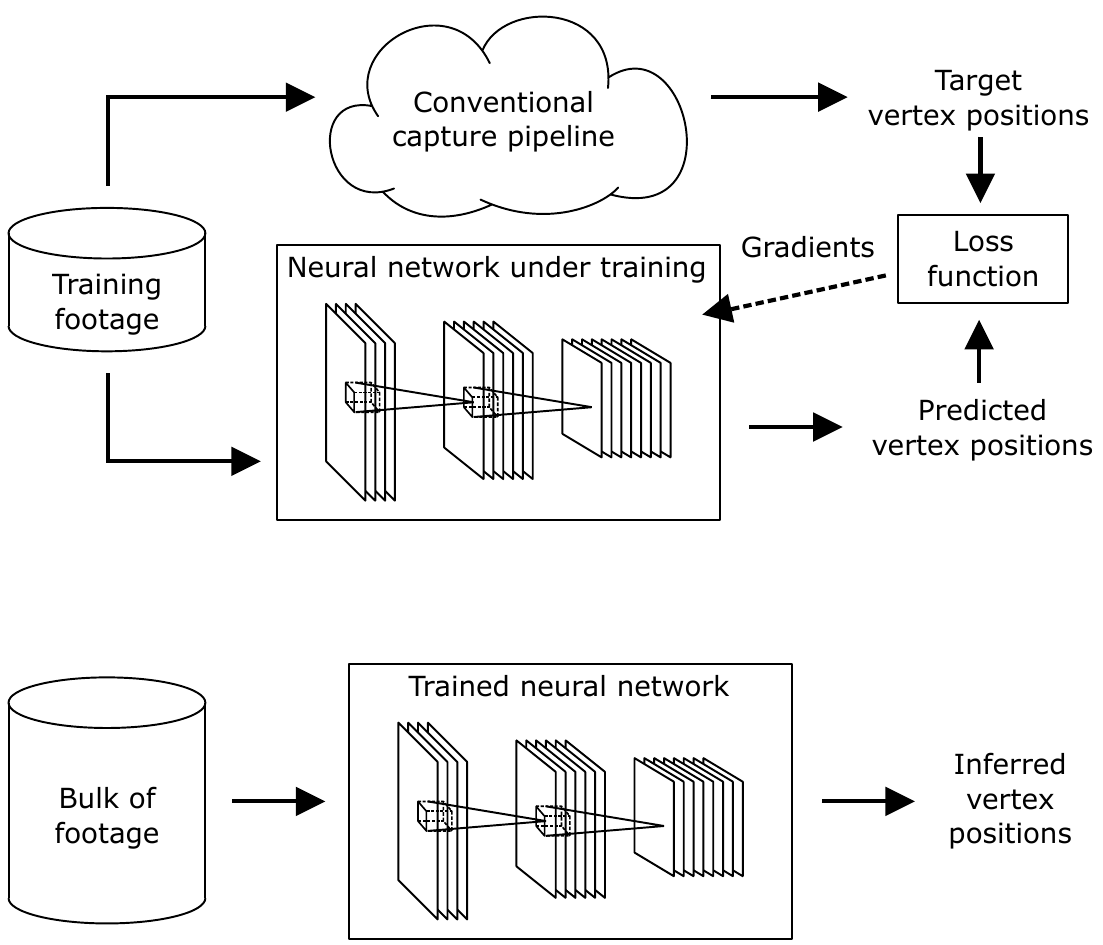}
\raisebox{31mm}[0mm]{Training}\\
\raisebox{2mm}[0mm]{Inference}
\caption{\label{fig:diagram}%
Our deep learning-based facial performance capture framework is divided into a training and inference stage.
The goal of our system is to reduce the amount of footage that needs to be
processed using labor-intensive production-level pipelines.
}
\end{figure}
}

\newcommand{\figprocess}{
\begin{figure}[t]
\centering
\begin{tabular}{@{}c@{\hspace*{8mm}}c@{}}
\includegraphics[width=0.43\linewidth]{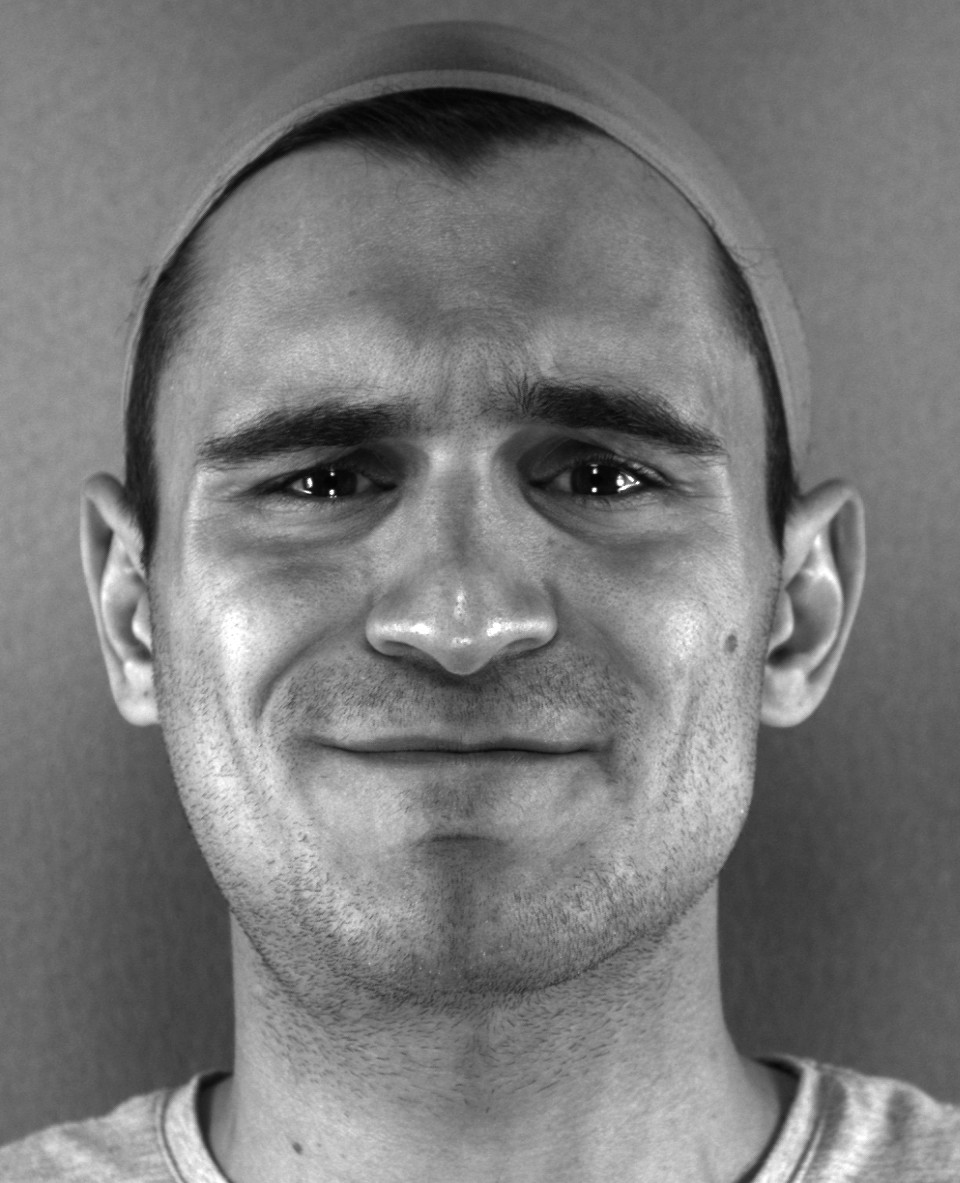}%
&\includegraphics[width=0.45\linewidth]{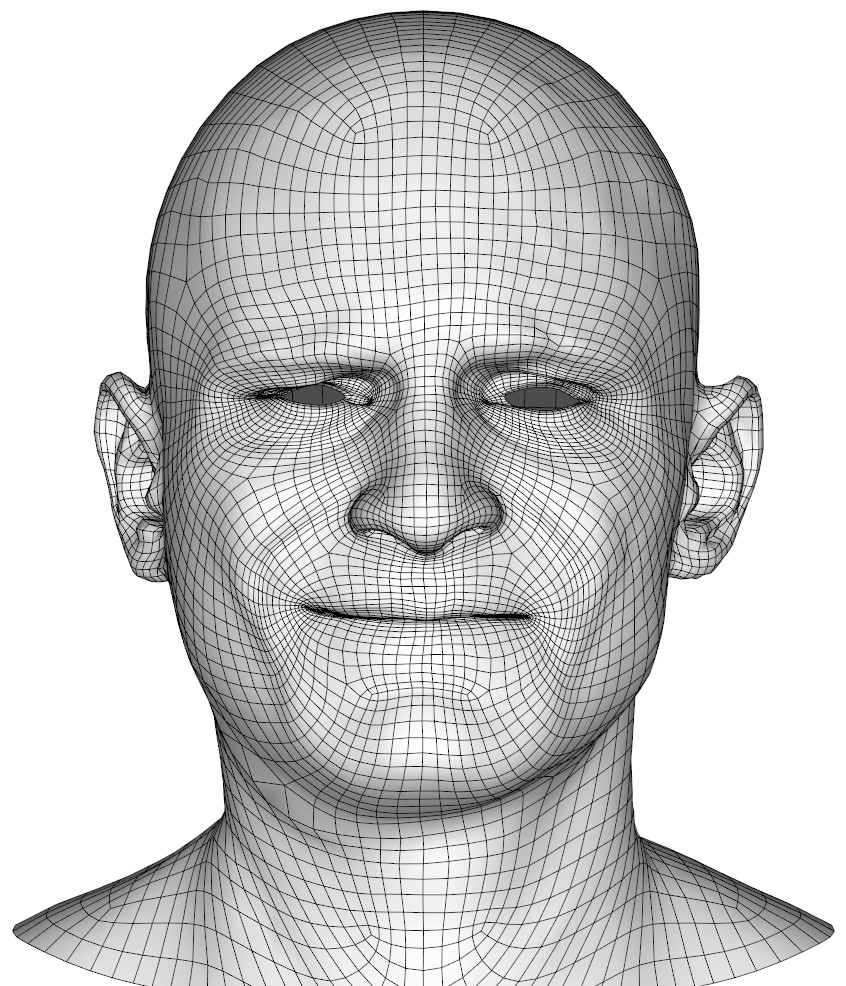}\\
Input video frame&Output%
\end{tabular}
\caption{\label{fig:process}%
Input for the conventional capture pipeline is a set of nine images, whereas our network only 
uses a cropped portion of the center camera image converted to grayscale.
Output of both the conventional capture pipeline and our network consists of  $\sim$5000 densely tracked 3D vertex positions for each frame.
}
\end{figure}
}

\newcommand{\figconvergence}{
\begin{figure}[t]
\centering
\vspace*{0.7mm}
\includegraphics[width=.997\linewidth]{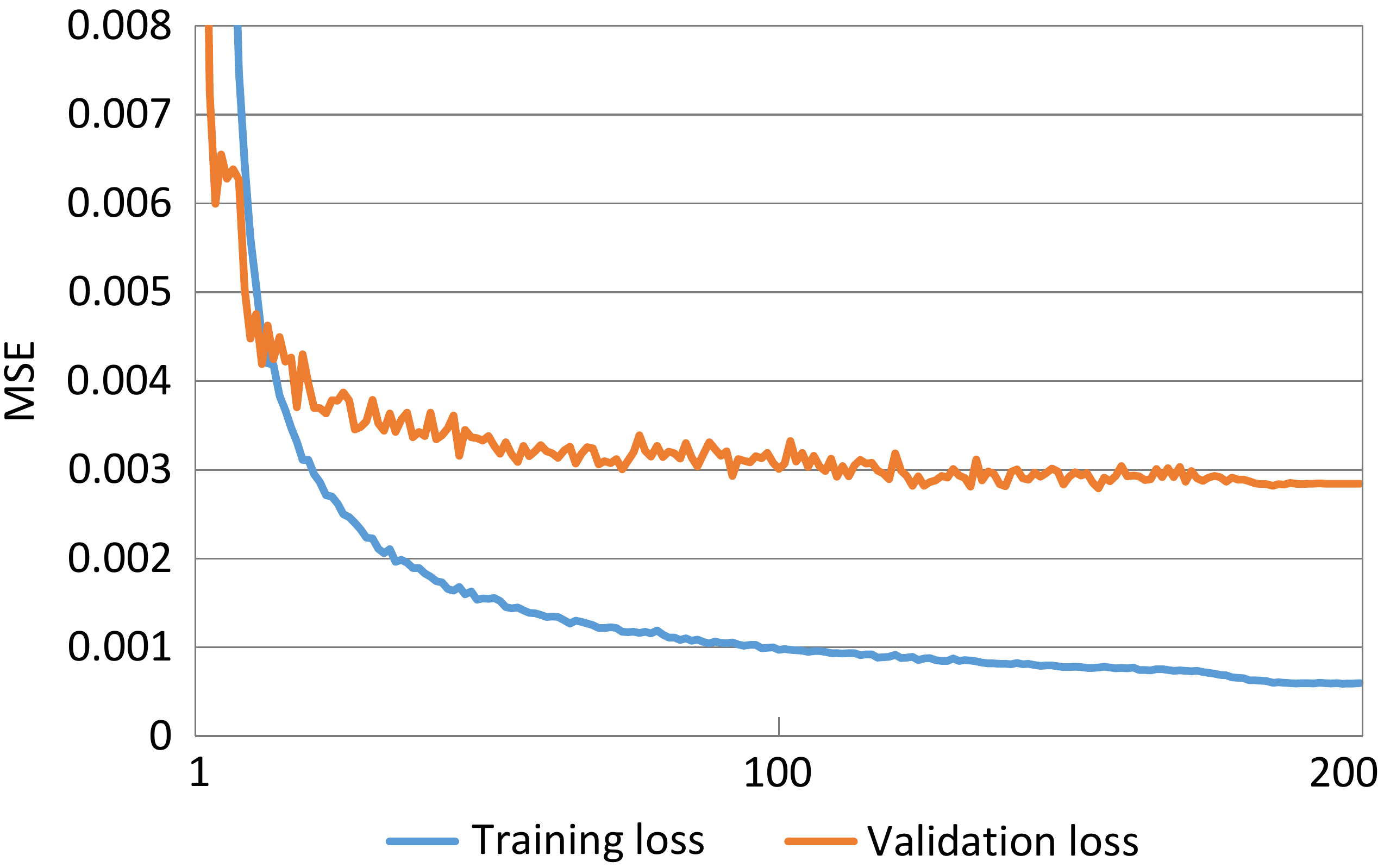}
\caption{\label{fig:convergence}%
Convergence of the convolutional network for Character~1 during 200 training epochs. The effects of learning
rate and $\beta_1$ rampdown can be seen in the final 30 epochs. This training run had 15173
training frames and 1806 validation frames and took 8 hours to finish. 
}
\end{figure}
}

\newcommand{\figconvergencefc}{
\begin{figure}[t]
\centering
\includegraphics[width=\linewidth]{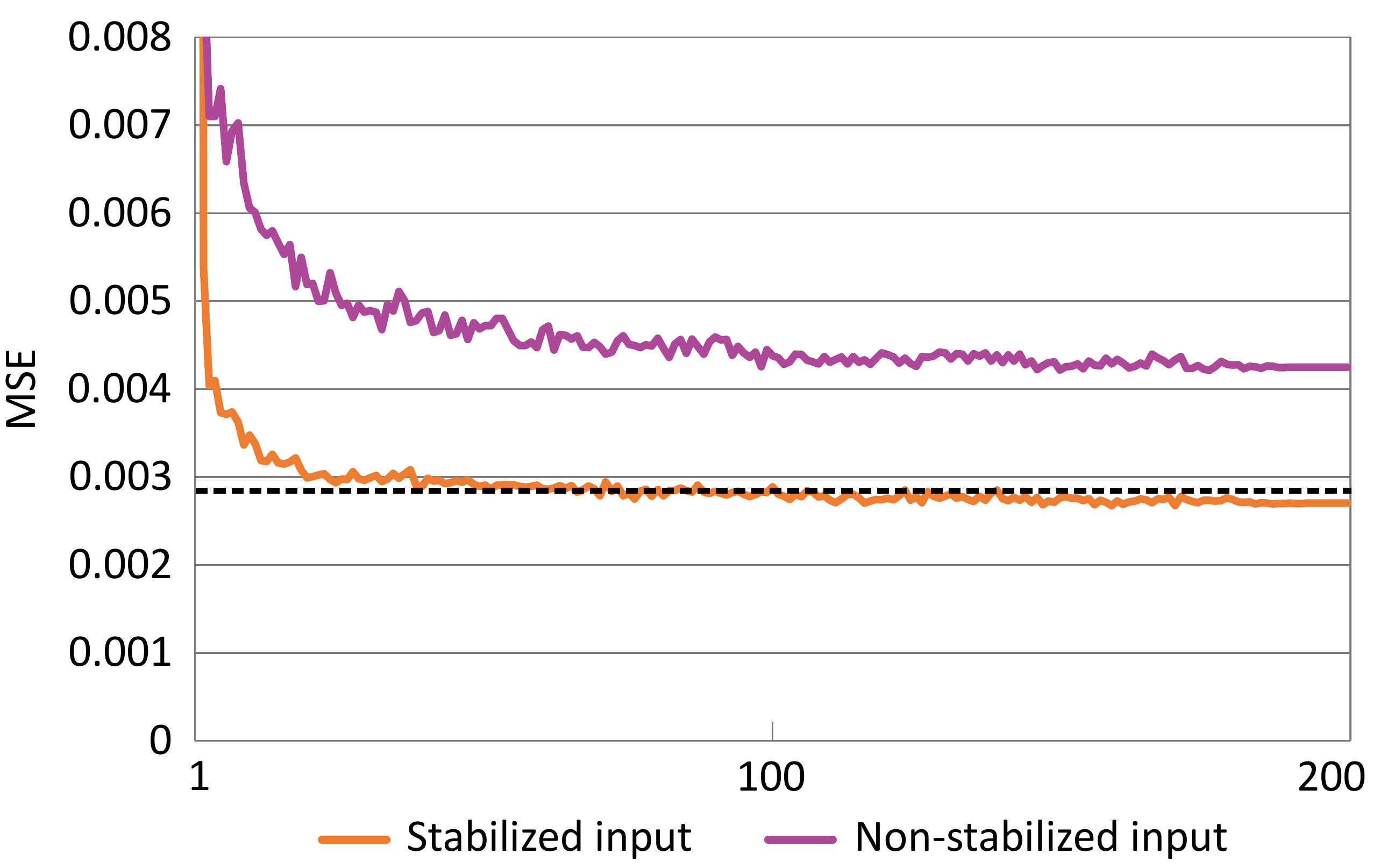}
\caption{\label{fig:convergencefc}%
Convergence of the validation loss of the fully-connected network. The training took approximately one hour,
and the training set was the same as for the convolutional network.
The dashed horizontal line shows the converged validation loss of the convolutional network, and we can see that with
stabilized input images, the fully connected network reaches slightly smaller loss than the all-convolutional network. However, visually the
results were clearly inferior. Using non-stabilized input images yielded much worse results both numerically
and visually.
}
\end{figure}
}

\newcommand{\rinput}[1] {\includegraphics[width=0.208\linewidth]{figures/results/input_#1.jpg}}
\newcommand{\rtarget}[1]{\includegraphics[width=0.218\linewidth]{figures/results/target_wire_#1.jpg}}
\newcommand{\routput}[1]{\includegraphics[width=0.218\linewidth]{figures/results/output_wire_#1.jpg}}
\newcommand{\rdiff}[1]  {\includegraphics[width=0.218\linewidth]{figures/results/diff_#1.jpg}}
\newcommand{\rrow}[1]{\rinput{#1}&\rtarget{#1}&\routput{#1}&\rdiff{#1}}
\newcommand{\rrowx}[2]{\rinput{#1}&\rtarget{#1}&\routput{#1}&\rdiff{#1}&\begin{sideways}\makebox[20mm]{\small RMSE=#2mm}\end{sideways}}
\newcommand{\rrowy}[3]{\rinput{#1}&\rtarget{#1}&\routput{#1}&\rdiff{#1}&\begin{sideways}\makebox[43mm]{\small RMSE = #2 / #3 mm}\end{sideways}}
\newcommand{\rheading}{(a) Input&(b) Reference &(c) Our&(d) Difference\\video frame&&result&\vspace*{1mm}}
\newcommand{\comparisonheading}{(a) Input video frame&(b) Reference &(c) Our result&(d) Difference\vspace*{1mm}}
\newcommand{\rbegintable}{\begin{tabular}{@{}cccc@{}c@{}}}
\newcommand{\scalebar}{\\{}\hfill\includegraphics[width=21mm]{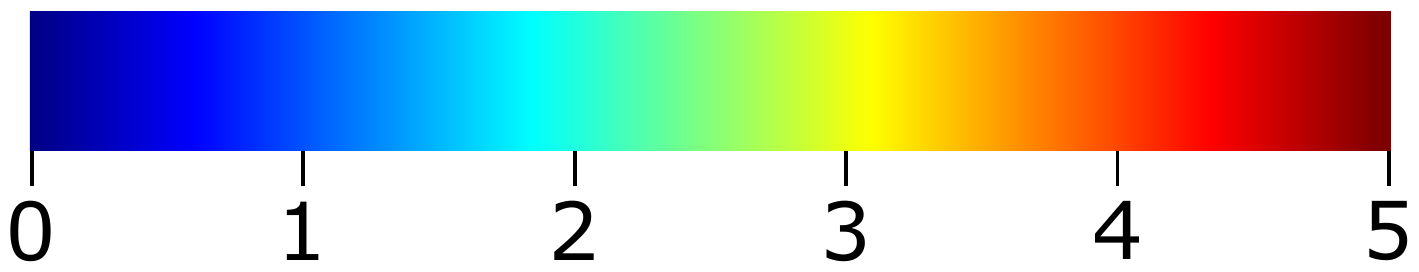}{\ \ }\\{}\vspace*{-.5mm}\hfill mm \hspace*{10.5mm}\vspace*{-1mm}}

\newcommand{\qinput}[1] {\includegraphics[width=0.208\linewidth]{figures/results2/input#1.jpg}}
\newcommand{\qtarget}[1]{\includegraphics[width=0.218\linewidth]{figures/results2/target#1.jpg}}
\newcommand{\qoutput}[1]{\includegraphics[width=0.218\linewidth]{figures/results2/output#1.jpg}}
\newcommand{\qdiff}[1]  {\includegraphics[width=0.218\linewidth]{figures/results2/diff#1.jpg}}
\newcommand{\qrow}[1]{\qinput{#1}&\qtarget{#1}&\qoutput{#1}&\qdiff{#1}}
\newcommand{\qrowx}[2]{\qinput{#1}&\qtarget{#1}&\qoutput{#1}&\qdiff{#1}&\begin{sideways}\makebox[20mm]{\small RMSE=#2mm}\end{sideways}}
\newcommand{\qrowy}[3]{\qinput{#1}&\qtarget{#1}&\qoutput{#1}&\qdiff{#1}&\begin{sideways}\makebox[43mm]{RMSE = #2 / #3 mm}\end{sideways}}

\newcommand{\figvalshotb}{
\begin{figure}[p]
\centering
\rbegintable
\rheading\\%
\rrowx{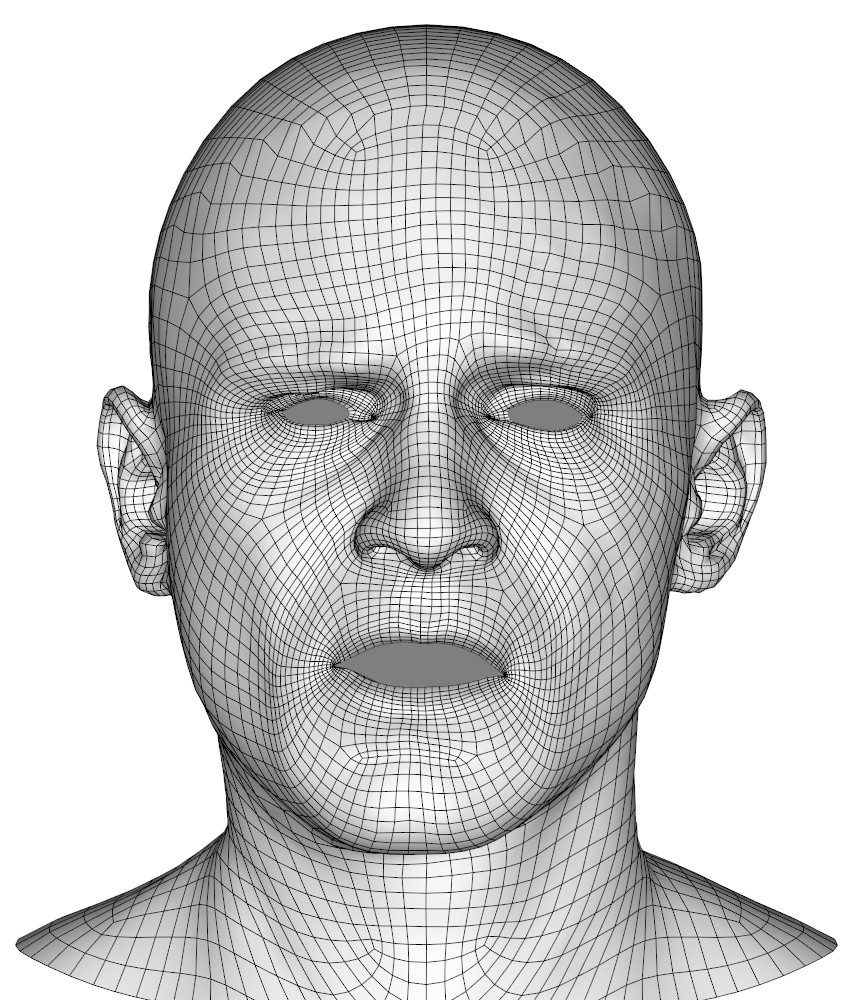}{1.18}\\
\rrowx{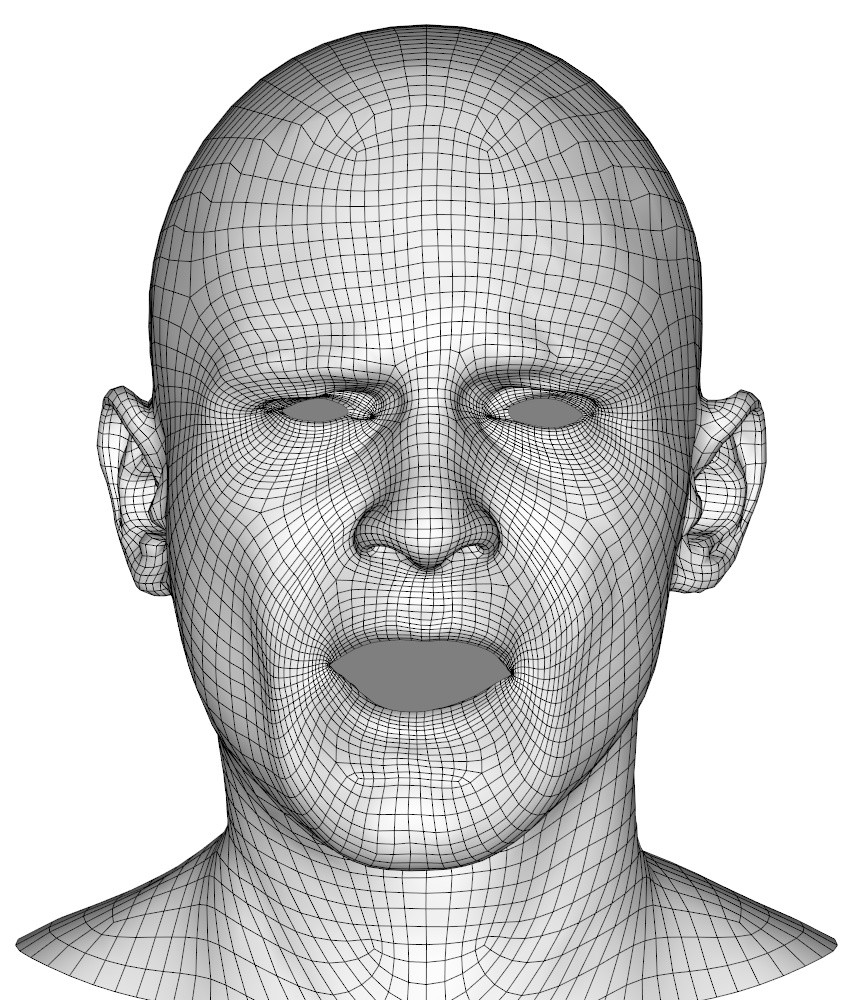}{1.37}\\
\rrowx{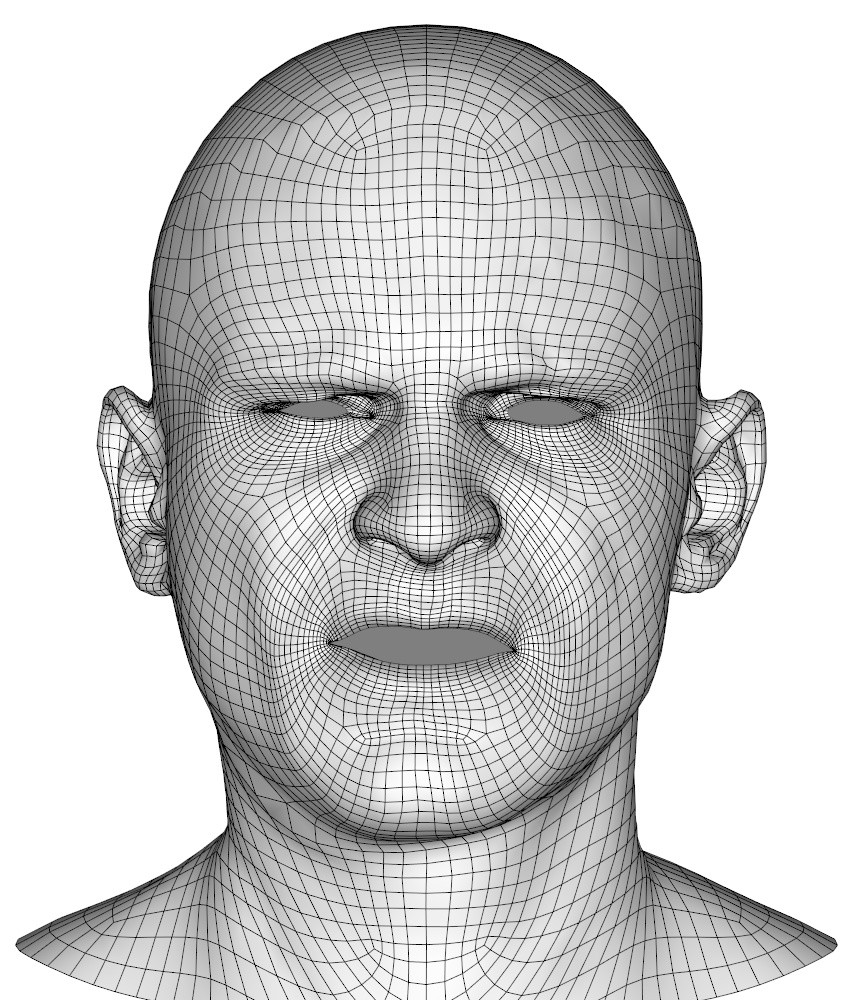}{1.19}\\
\rrowx{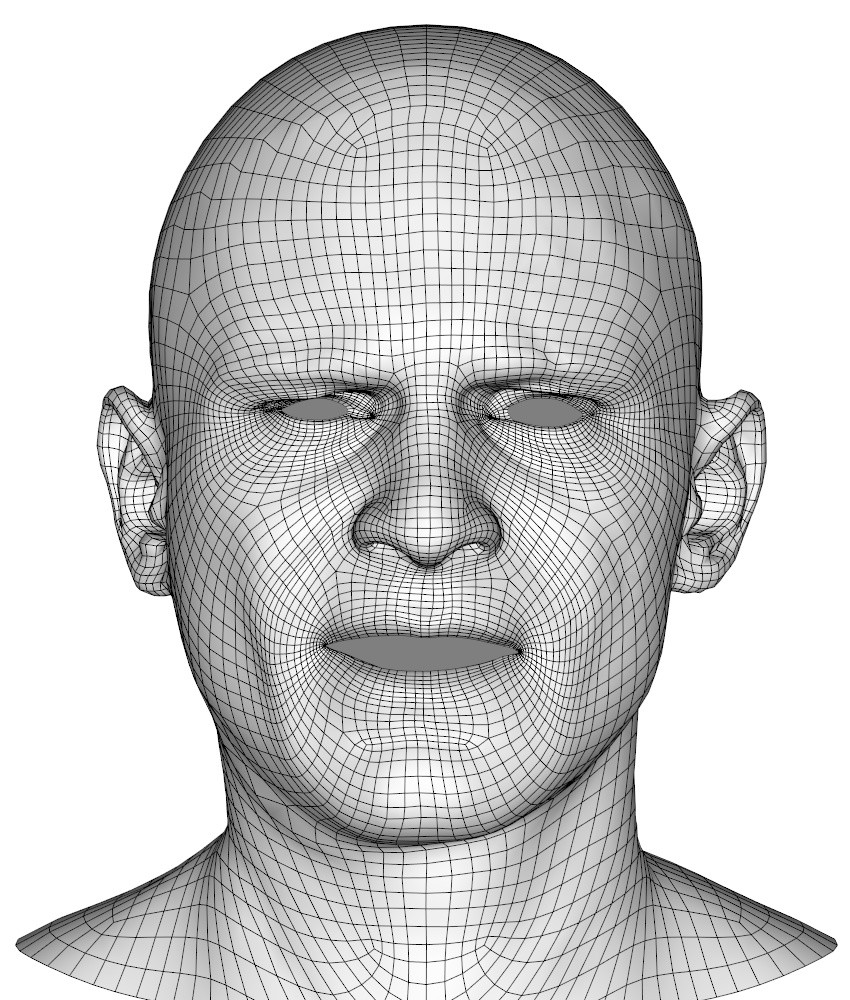}{1.00}\\
\qrowx{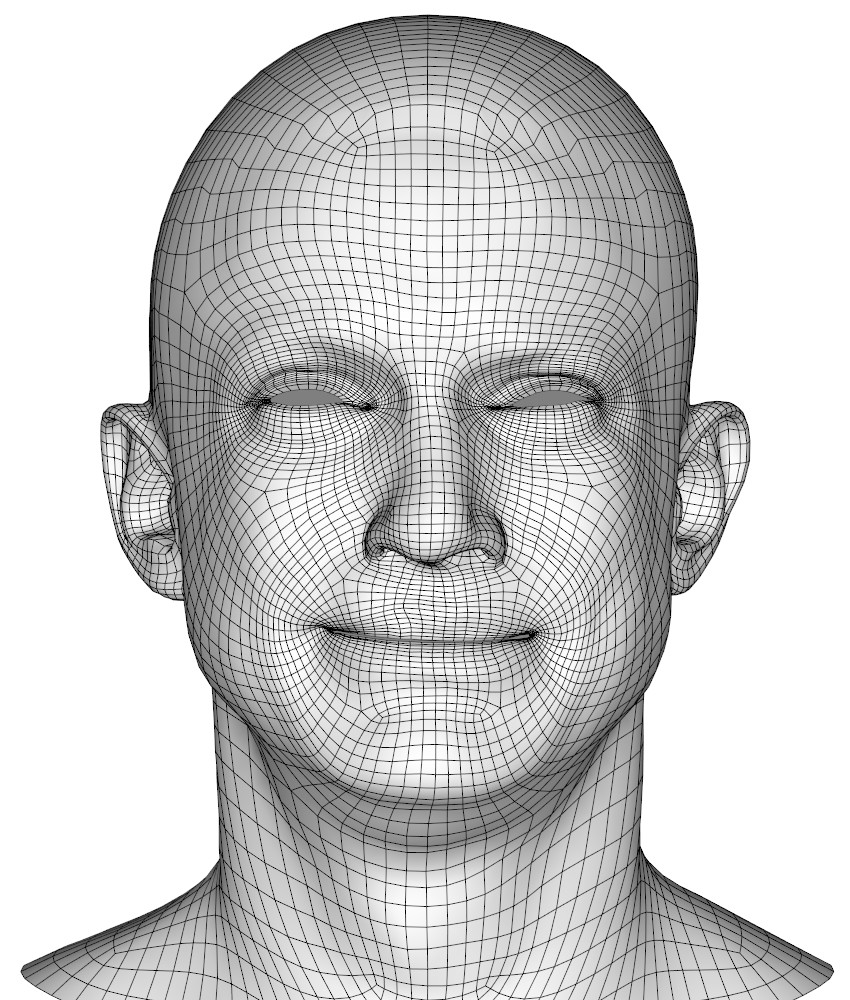}{1.27}\\ %
\qrowx{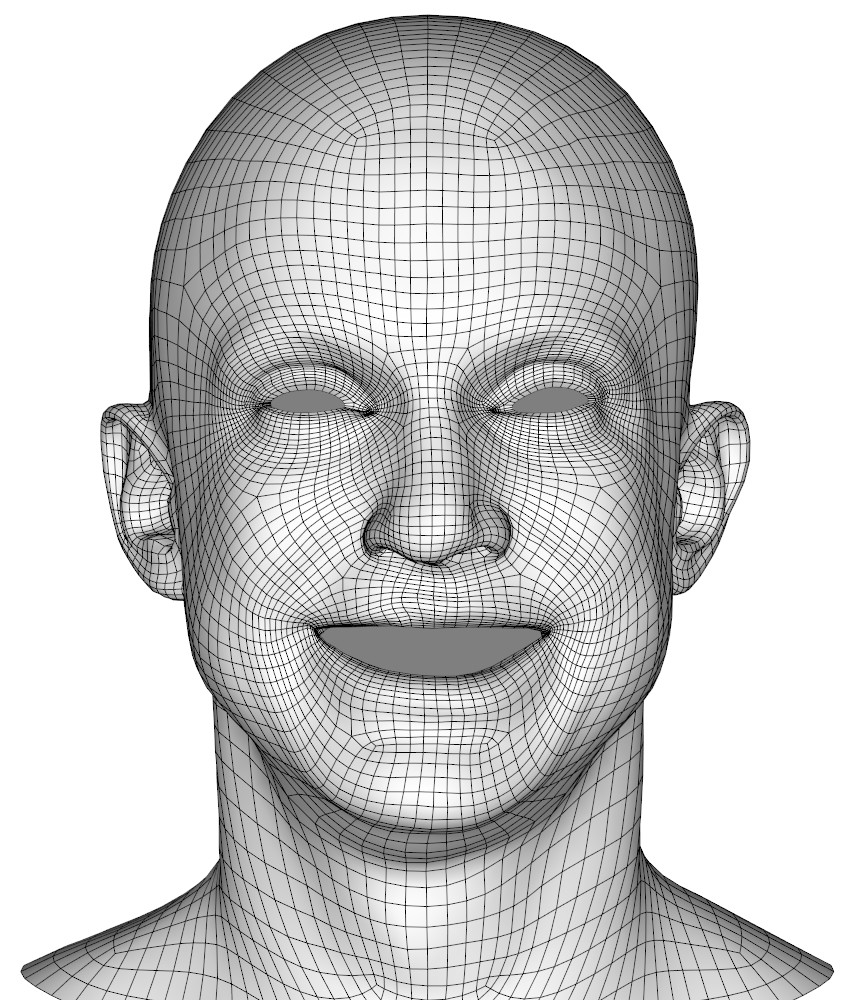}{1.28}\\ %
\qrowx{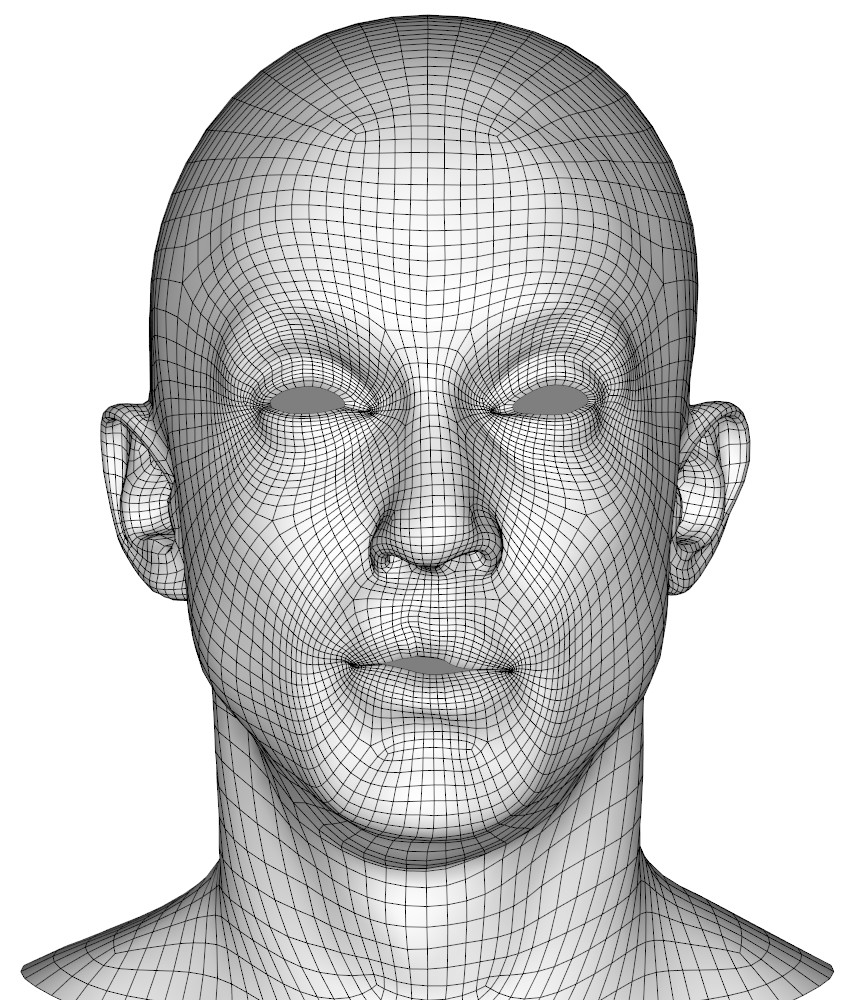}{0.70}\\ %
\end{tabular}
\scalebar
\caption{\label{fig:valshotb}%
A selection of frames from two validation shots.
(a) Crop of the original
input image. (b) The target positions created by capture artists using the existing
capture pipeline at Remedy Entertainment. (c) Our result inferred by the 
neural network based solely on the input image (a). (d) Difference between target and inferred positions. 
The RMSE is calculated over the Euclidean distances between target and inferred
vertex positions with only the animated vertices taken into account.
}
\vspace*{50mm}
\end{figure}
}

\newcommand{\figrmseplot}{
\begin{figure*}[t]
\centering
\includegraphics[width=\linewidth]{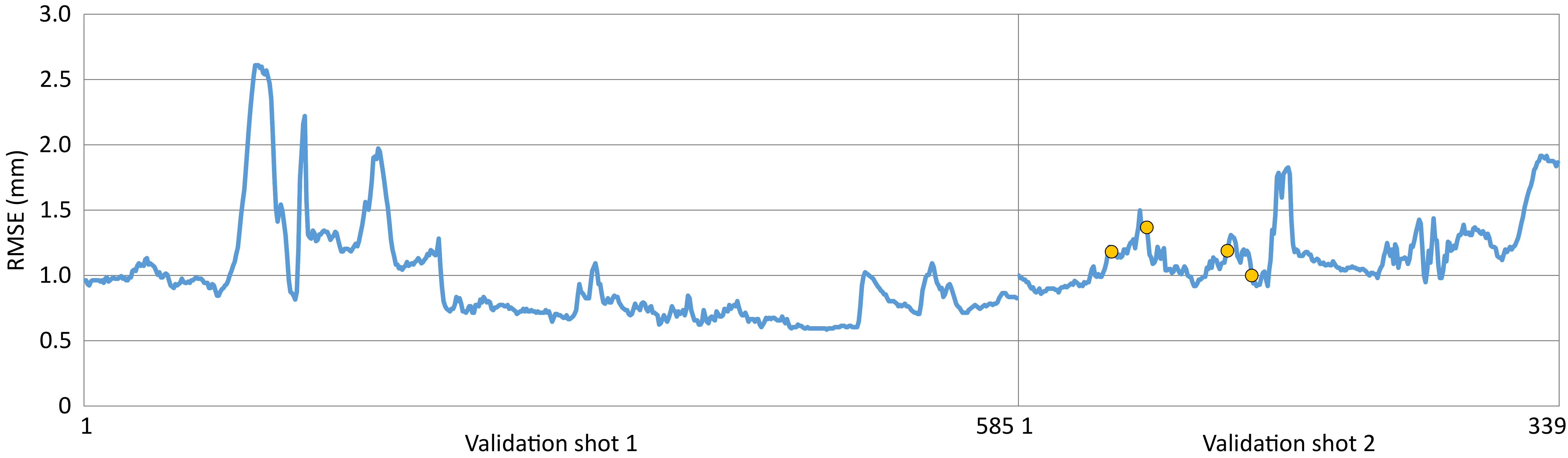}
\caption{\label{fig:rmseplot}%
Per-frame RMSE in the first two validation shots of Character~1.
Frame
index advances from left to right. 
The orange dots indicate frames used in the top four rows of Figure~\protect\ref{fig:valshotb}.
These validation shots are included in the accompanying video.
}
\end{figure*}
}

\newcommand{\augf}[1]{\includegraphics[width=0.242\linewidth]{figures/augment/aug#1.png}}
\newcommand{\augs}{\hspace*{0.01\linewidth}}
\newcommand{\figaugment}{
\begin{figure}[t]
\centering
\begin{tabular}{@{}c@{\augs}c@{\augs}c@{\augs}c@{}}
\augf{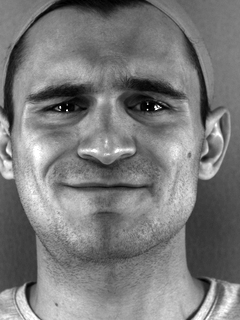}&\augf{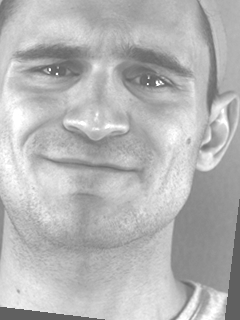}&\augf{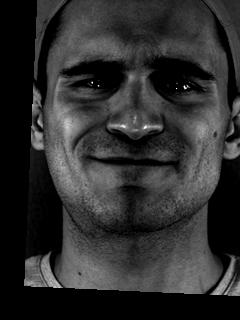}&\augf{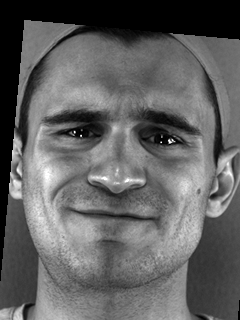}\vspace*{-0.2mm}\\
\augf{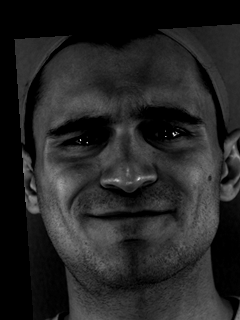}&\augf{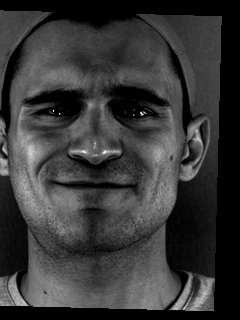}&\augf{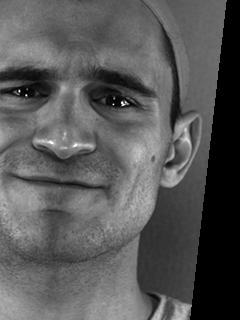}&\augf{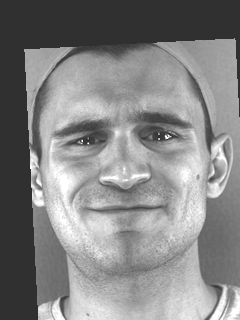}%
\end{tabular}
\caption{\label{fig:augment}%
Examples of augmented inputs presented to the network during training.
Top left image is the 240$\times$320 crop from the same input video frame as was
shown in Figure~\protect\ref{fig:process}, and the remaining images are augmented
variants of it.
}
\end{figure}
}

\newcommand{\ccrow}[1]{
\includegraphics[height=23.9mm]{figures/comp2/cinput-#1.jpg}&
\includegraphics[height=23.9mm]{figures/comp2/cao-#1.jpg}&
\includegraphics[height=23.9mm]{figures/comp2/thies-#1.jpg}&
\includegraphics[height=23.9mm]{figures/comp2/ours-#1.jpg}\\
}
\newcommand{\ccsp}{\hspace*{.5mm}}
\newcommand{\figcomparison}{
\begin{figure}[t]
\begin{tabular}{@{}c@{\ccsp}c@{\ccsp}c@{\ccsp}c@{}}
\centering
(a) Input &(b) Cao et &(c) Thies et&(d) Our\\ \vspace*{1mm}
video frame&al.~\shortcite{Cao2014}&al.~\shortcite{Thies2016}&result\\
\ccrow{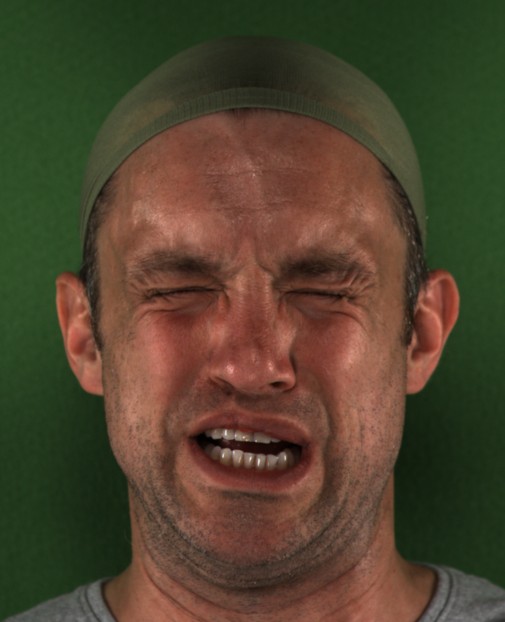}
\ccrow{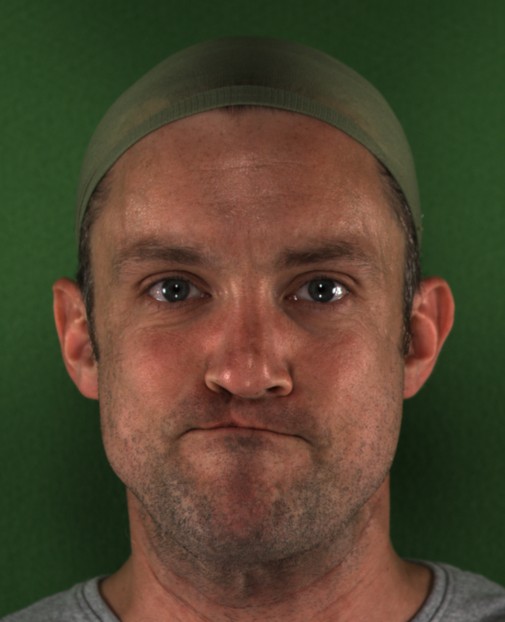}
\ccrow{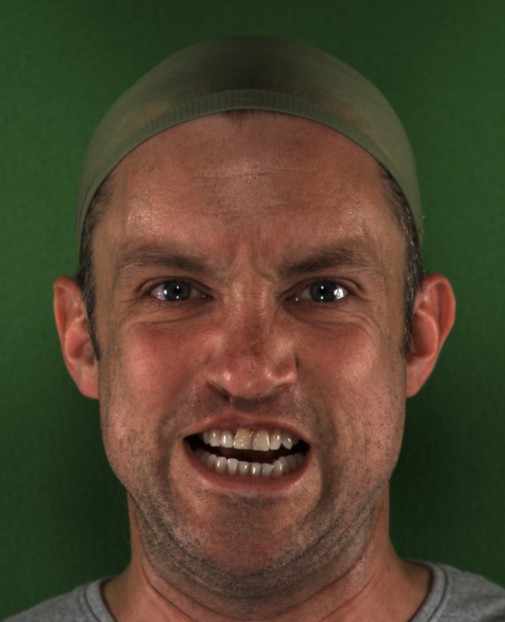}
\ccrow{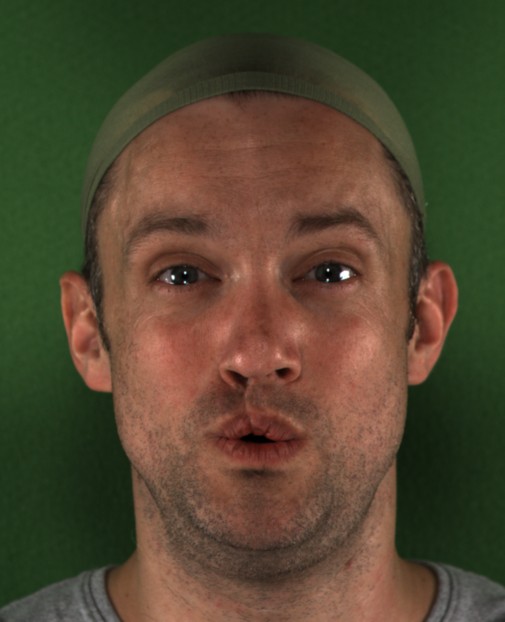}
\end{tabular}
\caption{We compare our method (d) with Thies et al. \shortcite{Thies2016} (c) and Cao et al. \shortcite{Cao2014} (b) on various input images (a). We see that our method is more accurate and able to better capture details around the difficult mouth and eye regions.}\label{fig:compare}
\end{figure}
}

\newlength{\tzerolen}
\newlength{\mzerolen}
\settowidth{\tzerolen}{0}
\settowidth{\mzerolen}{$0$}
\newcommand{\nout}{N_\mathrm{out}}
\newcommand{\x}{\hspace*{\tzerolen}}
\newcommand{\xx}{\hspace*{2\tzerolen}}
\newcommand{\z}{\hspace*{\mzerolen}}
\newcommand{\zz}{\hspace*{2\mzerolen}}
\renewcommand{\paragraph}[1]{\emph{#1} } %

\begin{abstract}
We present a real-time deep learning framework for video-based facial performance capture---the dense 3D tracking of an actor's face given a monocular video. 
Our pipeline begins with accurately capturing a subject using a high-end production facial capture pipeline based on multi-view stereo
tracking and artist-enhanced animations. 
With 5--10 minutes of captured footage, we train a convolutional neural network to produce high-quality output, including self-occluded regions, from a monocular video sequence of that subject.
Since this 3D facial performance capture is fully automated, our system can drastically reduce the amount of labor involved in the development
of modern narrative-driven video games or films involving realistic digital doubles of actors and potentially hours of animated dialogue per character.
We compare our results with several state-of-the-art monocular real-time facial capture techniques and demonstrate compelling animation inference in challenging 
areas such as eyes and lips.
\end{abstract}

\ccsdesc[500]{Computing methodologies~Animation}
\ccsdesc[500]{Computing methodologies~Neural networks}
\ccsdesc[300]{Computing methodologies~Supervised learning by regression}

\keywords{Facial animation, deep learning}

\maketitle
\section{Introduction}
\label{sec:intro}

\figdiagram

The use of visually compelling digital doubles of human actors is a key component for increasing realism 
in any modern narrative-driven video game. Facial performance capture poses many challenges in computer 
animation and due to a human's innate sensitivity to the slightest
facial cues, it is difficult to surpass the uncanny valley, where otherwise believable 
renderings of a character appear lifeless or unnatural.

Despite dramatic advancements in automated facial performance capture systems and their wide deployment for scalable production, it is still not possible
to obtain a perfect tracking for highly complex expressions, especially in challenging but critical areas such as lips and eye regions. In most cases,
manual clean-up and corrections by skilled artists are necessary to ensure high-quality output that is free from artifacts and noise.
Conventional facial animation pipelines can easily result in drastic costs, especially in 
settings such as video game production where hours of footage may need to be processed.

In this paper, we introduce a deep learning framework for real-time and production-quality facial performance capture.
Our goal is not to fully eliminate the need for manual work, but to significantly reduce the extent to which it is required. 
We apply an offline, multi-view stereo capture pipeline with manual clean-up 
to a small subset of the input video footage, 
and use it to generate enough data to train a deep neural network. The trained network can then be used to automatically process the remaining video footage at rates as fast as 870 fps, 
skipping the conventional labor-intensive capture pipeline entirely. 

Furthermore, we only require a single view as input during runtime which makes our solution attractive for head cam-based 
facial capture.
Our approach is real-time and does not even need sequential processing, so every frame can be processed independently.
Furthermore, we demonstrate qualitatively superior results compared to state-of-the-art monocular real-time facial capture solutions.
Our pipeline is outlined in Figure~\ref{fig:diagram}.

\subsection{Problem Statement}

We assume that the input for the capture pipeline is multiple-view videos of the actor's head captured under controlled conditions to generate training data for the neural network.
The input to the neural network at runtime is video from a single view.
The
positions of the cameras remain fixed, the lighting and background are standardized, and the actor is to remain at approximately the
same position relative to the cameras throughout the recording. Naturally, some amount of movement needs to be allowed, and we achieve this
through data augmentation during training (Section~\ref{sec:augment}).

The output of the capture pipeline is the set of per-frame positions of facial mesh vertices, as illustrated in Figure~\ref{fig:process}. 
Other face encodings such as blendshape weights or joint positions are introduced in later stages of our pipeline, mainly for compression and rendering purposes, but the primary capture output
consists of the positions of approximately 5000 animated vertices on a fixed-topology facial mesh.

\subsection{Offline Capture Pipeline}
The training data used for the deep neural network was generated using Remedy Entertainment's in-house capture pipeline
based on a cutting edge commercial DI4D PRO system \cite{di4d} that employs nine video cameras.

\figprocess

First, an unstructured mesh with texture and optical flow data is created from the images for each frame of an input video.
A fixed-topology template mesh is created prior to the capture work by applying Agisoft \cite{agisoft}, a standard multi-view stereo reconstruction software, on data from 26 DSLR cameras and two cross polarized flashes. 
The mesh is then warped onto the unstructured scan of the first frame. 
The template
mesh is tracked using optical flow through the entire sequence. Possible artifacts are manually fixed using the DI4DTrack software by a clean-up artist.
The position and orientation of the head are then stabilized using a few key vertices of the tracking mesh. The system then outputs the positions of each of the vertices on the fixed-topology template mesh.

Additional automated deformations are later applied to the vertices to fix remaining issues. 
For instance, the eyelids are deformed to meet the eyeballs exactly and to slide slightly with motion of
the eyes. Also, opposite vertices of the lips are smoothly brought together to improve lip contacts when needed. 
After animating the eye directions the results are compressed for runtime use in Remedy's Northlight engine using 416 facial joints. 
Pose space deformation is used to augment the facial animation with detailed wrinkle normal map blending.
These ad-hoc deformations were not applied in the training set. 

\section{Related Work}
\label{sec:prevwork}

The automatic capture of facial performances has been an active area of research for decades \cite{williams1990,Blanz1999,Pughin2006,Mattheyses2015}, and is widely used in game and movie production today.
In this work we are primarily interested in real-time methods that are able to track the entire face, without relying on markers, and are based on consumer hardware, ideally a single RGB video camera.

\subsection{Production Facial Capture Systems}
Classic high-quality facial capture methods for production settings require markers \cite{williams:90, Guenter:1998:MF, Bickel:2008:PAT} or other application specific hardware  \cite{PighinLewisCourse06}. 
Purely data-driven high-quality  facial capture methods used in a production setting still require complicated hardware and camera setups \cite{Borshukov:2005:UCI, Vlasic:2005:FTM, Alexander:2009:DEP, zhang2004,Weise2009,beeler2011,Bhat2013,Fyffe2014}
and a considerable amount of computation such as multi-view stereo or photometric reconstruction of individual input frames \cite{Bradley:2010:HRP, Valgaerts:2012:LBF, furukawa2009,beeler2011,fyffe2011,Shi2014}
that often require carefully crafted 3D tracking model \cite{Borshukov:2005:UCI,Vlasic:2005:FTM, Alexander:2009:DEP}.
Many production setting facial performance capture techniques require extensive manual post-processing as well.

Specialized techniques have also been proposed for various subdomains, including eyelids \cite{Bermano2015}, gaze direction \cite{zhang15}, lips \cite{Garrido2016}, handling of occlusions \cite{Saito2016}, and handling of extreme skin deformations \cite{Wu2016}.

Our method is a ``meta-algorithm'' in the sense that it relies on an existing technique for generating the training examples, and then learns to mimic the host algorithm, producing further results at a fraction of the cost. As opposed to the complex hardware setup, heavy computational time, and extensive manual post-processing involved in these production setting techniques, our method is able to produce results with a single camera, reduced amounts of manual labor, and at a rate of slightly more than 1 ms per frame when images are processed in parallel.
While we currently base our system on a specific commercial solution, the same general idea can be built on top of any facial capture technique taking video inputs, ideally the highest-quality solution available.

\subsection{Single-View Real-time Facial Animation}

Real-time tracking from monocular RGB videos is typically based either on detecting landmarks and using them to drive the facial expressions \cite{Cootes2001, Saragih:2011:DMF, Tresadern2012} or on 3D head shape regression \cite{hsieh2015unconstrained,  weise2009face, li2010example,  Cao2013,Cao2014,Cao2015,Thies2016,Olszewski2016}. 
Of these methods, the regression approach has delivered higher-fidelity results, and real time performance has been demonstrated even on mobile devices \cite{Weng2014}. The early work on this area \cite{Cao2013,Weng2014} require an actor-specific training step, but later developments have relaxed that requirement \cite{Cao2014} and also extended the method to smaller-scale features such as wrinkles \cite{Cao2015,Ichim2015}. 

Most of these methods are targeting ``in-the-wild" usage and thus have to deal with varying lighting, occlusions, and unconstrained head poses. Thus, these methods are typically lower quality in detail and accuracy.
These methods are also usually only able to infer low-dimensional facial expressions---typically only a few blendshapes---reliably. 
More problems also arise in appearance based methods such as \cite{Thies2016}.
For example, relying on pixel constraints makes it possible only to track visible regions, making it difficult to reproduce regions with complex interactions such as the eyes and lips accurately.
Additionally, relying on appearance can lead to suboptimal results if the PCA model does not accurately encode the subject's appearance such as in the case of facial hair.

In contrast, we constrain the setup considerably in favor of high-fidelity results for one particular actor. In our setup, all of the lighting and shading as well as gaze direction and head poses are produced at runtime using higher-level procedural controls.
Using such  a setup, unlike the other less constrained real-time regression-based methods, our method is able obtain high quality results as well as plausible inferences for the non-visible regions and other difficult to track regions such as the lips and eyes.

Olszewski et al.~\shortcite{Olszewski2016} use neural networks to regress eye and mouth videos separately into blend shape weights in a head-mounted display setup. Their approach is closely related to ours with some slight differences. First of all, their method considers the eye and mouth separately while our method considers the whole face at once.  Also, they use blendshapes from FACS \cite{Ekman:1978:FAC} while our system 
produces vertex coordinates of the face mesh based on a 160-dimensional PCA basis. Moreover, our system can only process one user at a time without retraining while the method of Olszewski et al.~\shortcite{Olszewski2016} is capable of processing several different identities. However, our method can ensure accurate face tracking while theirs is only designed to track the face to drive a target character.

\subsection{Alternative Input Modalities}
Alternatively, a host of techniques exists for audio-driven facial animation \cite{Cohen1993,Brand1999,Taylor2012,JALI}, and while impressive results have been demonstrated, these techniques are obviously not applicable to non-vocal acting and also commonly require an animator to adjust the correct emotional state. They continue to have important uses as a lower-cost alternative, e.g., in in-game dialogue. 

A lot of work has also been done for RGB-D sensors, such as Microsoft Kinect Fusion, e.g.,~\cite{Weise2011,Li2013,Bouaziz2013,Hsieh2015,Thies2015}.
Recently Liu et al. also described a method that relies on RGB-D and audio inputs \cite{Liu2015}.

\subsection{Convolutional Neural Networks (CNN)}
We base our work on deep CNNs that have received significant attention in the recent years, and proven particularly well suited for large-scale image recognition tasks \cite{krizhevsky2012,simonyan2014}. Modern CNNs employ various techniques to reduce the training time and improve generalization over novel input data, including data augmentation \cite{simard2003}, dropout regularization \cite{srivastava2014}, ReLU activation functions, i.e., $max(0,\cdot)$, and GPU acceleration \cite{krizhevsky2012}. Furthermore, it has been shown that state-of-the-art performance can be achieved with very simple network architectures that consist of small 3$\times$3-pixel convolutional layers \cite{simonyan2014} that employ strided output to reduce spatial resolution throughout the network \cite{springenberg2014}.

\section{Network Architecture}
\label{sec:network}

Our input footage is divided into a number of shots, with each shot typically consisting of 100--2000 frames at 30 FPS. Data for each input frame
consists of a 1200$\times$1600 pixel image from each of the nine cameras. As explained above, the output is the per-frame vertex position
for each of the $\sim$5000 facial mesh vertices. %

As input for the network, we take the 1200$\times$1600 video frame from the central camera, crop it with a fixed rectangle so that
the face remains in the picture, and scale the remaining portion to 240$\times$320 resolution. Furthermore, we convert the image to
grayscale, resulting in a total of 76800 scalars to be fed to the network. The resolution may seem low, but numerous tests confirmed that
increasing it did not improve the results.

\subsection{Convolutional Network}
\label{sec:convnet}

Our convolutional network is based on the all-convolutional architecture \cite{springenberg2014} extended with two fully connected layers
to produce the full-resolution vertex data at output.
The input is a whitened version of the 240$\times$320 grayscale image. For whitening, we calculate the mean and variance over all pixels
in the training images, and bias and scale the input so that these are normalized to zero and one, respectively.

Note that the same whitening coefficients, fixed at training time, are used for all input images during training, validation, and production use.
If the whitening were done on a per-image or per-shot basis, we would lose part of the benefits of the standardized lighting environment.
For example, variation in the color of the actor's shirt between shots would end up affecting the brightness of the face.
The layers of the network are listed in the table below.

\newcommand{\npad}{\hspace*{2.5mm}}
\newcommand{\dpad}{\hspace*{48mm}}
\newcommand{\vpad}{\vspace*{2mm}}
\newcommand{\rbh}[1]{\raisebox{-0.25mm}{#1}}

\vpad
\begin{center}
\begin{tabular}{|l|l|}
\hline
\rbh{Name}\npad{} & \rbh{Description}\dpad{} \\
\hline
\raisebox{0mm}[3mm]{}%
input  & Input $1\times240\times320$ image \\
conv1a & Conv $3\times3$, $\zz1\to\z64$, stride $2\times2$, ReLU \\
conv1b & Conv $3\times3$, $\z64\to\z64$, stride $1\times1$, ReLU \\
conv2a & Conv $3\times3$, $\z64\to\z96$, stride $2\times2$, ReLU \\
conv2b & Conv $3\times3$, $\z96\to\z96$, stride $1\times1$, ReLU \\
conv3a & Conv $3\times3$, $\z96\to 144$, stride $2\times2$, ReLU \\
conv3b & Conv $3\times3$, $ 144\to 144$, stride $1\times1$, ReLU \\
conv4a & Conv $3\times3$, $ 144\to 216$, stride $2\times2$, ReLU \\
conv4b & Conv $3\times3$, $ 216\to 216$, stride $1\times1$, ReLU \\
conv5a & Conv $3\times3$, $ 216\to 324$, stride $2\times2$, ReLU \\
conv5b & Conv $3\times3$, $ 324\to 324$, stride $1\times1$, ReLU \\
conv6a & Conv $3\times3$, $ 324\to 486$, stride $2\times2$, ReLU \\
conv6b & Conv $3\times3$, $ 486\to 486$, stride $1\times1$, ReLU \\
drop   & Dropout, $p=0.2$ \\
fc     & Fully connected $9720\to160{}$, linear activation \\
output & Fully connected $160\to\nout{}$, linear activation \\
\hline
\end{tabular}
\end{center}
\vpad

The output layer is initialized by precomputing a PCA basis for the output meshes based on the target meshes
from the training data. Allowing 160 basis vectors explains approximately 99.9\% of the variance seen in the meshes, which was
considered to be sufficient. If we 
fixed the weights of the output layer and did not train them,
that would effectively
train the remainder of the network to output the 160 PCA coefficients. However, we found that allowing the last layer to be trainable as well improved the results.
\FINAL{This would seem to suggest that the optimization is able to find a slightly better intermediate basis than the initial PCA basis.}

\section{Training}
\label{sec:training}

For each actor, the training set consists of four parts, totaling approximately 5--10 minutes
of footage. The composition of the training set is as follows.

\paragraph{Extreme Expressions.} In order to capture the maximal extents of the facial motion, a single
range-of-motion shot is taken where the actor goes through a pre-defined set of extreme expressions. 
These include but are not limited to opening the mouth as wide as possible, moving the jaw
sideways and front as far as possible, pursing the lips, and opening the eyes wide and forcing them shut.

\paragraph{FACS-Like Expressions.} Unlike the range-of-motion shot that contains exaggerated expressions, this
set contains regular FACS-like expressions such as squinting of the eyes or an expression of disgust.
These kind of expressions must be included in the training set as otherwise the network would not
be able to replicate them in production use.

\paragraph{Pangrams.} This set attempts to cover the set of possible facial motions during normal speech for a
given target language, in our case English. The actor speaks one to three pangrams, which are sentences that are designed
to contain as many different phonemes as possible, in several different emotional tones. A pangram
fitting the emotion would be optimal but in practice this is not always feasible.

\paragraph{In-Character Material.} This set leverages the fact that an actor's performance of a character
is often heavily biased in terms of emotional and expressive range for various dramatic and narrative reasons.
This material is composed of the preliminary version of the script, or it may be otherwise prepared for the
training. Only the shots that are deemed to support the different aspects of the character are selected so as
to ensure that the trained network produces output that stays in character even if the inference isn't
perfect or if completely novel or out of character acting is encountered.
 
The training set is typically comprised of roughly 10\% of range-of-motion and expression shots, 30\% of pangrams across emotional states, and 60\% of in-character performances of varying intensity and scenario. %

\subsection{Data Augmentation}
\label{sec:augment}

We perform several transformations to the input images during training in order to make the network resistant to
variations in input data. These transformations are executed on CPU concurrently with network
evaluation and training that occurs on the GPU. Augmentation is not used when evaluating the validation loss
or when processing unseen input data in production use.
Examples of augmented input images are shown in Figure~\ref{fig:augment}.

The main transformations are translation, rotation and zoom, which account for the motion of the
actor's head during capture. The magnitudes of these augmentations are set so that they cover at least
all of the variation expected in the input data.
\FINAL{This kind of image-based augmentation does not cover large-scale changes in head pose, and thus our method does not tolerate that unless such effects are present in the training data.}

In addition to geometric transformations, we vary the brightness and contrast of the input images
during training, in order to account for variations in lighting over the capture process. Our
cameras pick a slight periodic flicker from the 50~Hz LED lights in the capture room, and it is possible that
some of the bulbs degrade during the capture period that may take place over several days or weeks.

\figaugment

\subsection{Training Parameters}

We train the network for 200 epochs using the Adam \cite{kingma2014} optimization algorithm with parameters
set to values recommended in the paper. The learning rate is ramped up using a geometric progression during
the first training epoch, and then decreased according to \raisebox{0mm}[0mm]{\mbox{$1/\sqrt{t}$}} schedule. During the last~30
epochs we ramp the learning rate down to zero using a smooth curve, and simultaneously ramp Adam $\beta_1$ parameter from~\mbox{0.9}
to~\mbox{0.5}. The ramp-up removes an occasional glitch where the network does not start learning at all, 
and the ramp-down ensures that the network converges to a local minimum. Minibatch size is set to 50, and 
each epoch processes all training frames in randomized order. Weights are initialized using the
initialization scheme of He~et~al.~\shortcite{he2015}, except for the last output layer which is initialized
using a PCA transformation matrix as explained in Section~\ref{sec:convnet}.

The strength of all augmentation transformations is ramped up linearly during the first five training
epochs, starting from zero. This prevented a rare but annoying effect where the network fails
to start learning properly, and gets stuck at clearly sub-optimal local minimum.
The augmentation ramp-up process can be seen as a form of curriculum learning \cite{bengio2009}.

Our loss function is simply the mean square error between the predicted vertex positions produced by the network
and the target vertex positions in the training data. 

Our implementation is written in Python using Theano \cite{theano} and Lasagne \cite{Lasagne}.
On a computer with a modern CPU and a NVIDIA Titan X GPU, the training of one network with a typical
training set containing 10000--18000 training frames ($\sim$5--10 minutes at 30Hz) takes approximately~5--10 hours.

\section{Results}
\label{sec:results}

\FINAL{We tested the trained network using footage from a later session.} The
lighting conditions and facial features exhibited in the training set were carefully preserved.
The inference was evaluated numerically and perceptually in relation to a manually tracked ground
truth. 

We will first evaluate our choices in the design and training of our neural network, followed by
examination of the numerical results. We then turn to visual comparisons with recent monocular real-time facial performance capture methods. 
Finally, we explore the limitations of our pipeline.

The quality of the results can be best assessed from the accompanying video. In the video an interesting observation is that our results are not only accurate but also perfectly stable temporally despite the fact that we do not employ recurrent networks or smooth the generated vertex positions temporally in any way.
It is very difficult for human
operators to achieve similar temporal stability as they inevitably vary in their work between sequences.

\subsection{Network Architecture Evaluation}
\label{sec:fcnet}

All results in this paper were computed using the architecture described in Section ~\ref{sec:convnet}.
It should be noted that the quality of the results is not overly sensitive to the exact composition of the network.
Changing the dimensions of the convolutional layers or removing or adding the \mbox{$1\times1$}
stride convolution layers only changed performance by a slight margin. The architecture described in Section ~\ref{sec:convnet}  was found to perform slightly better compared to other all-convolutional architectures that could be trained in a reasonable amount of time, so it was chosen for
use in production.

In addition to using an all-convolutional neural network, we also experimented with fully connected networks.
When experimenting with fully connected networks, we achieved the best results by transforming the input images into 3000 PCA
coefficients. The PCA basis is pre-calculated based on the input frames from the training set, and the chosen
number of basis images captures approximately 99.9\% of the variance in the data. The layers of the network
are listed in the table below.

\vpad%
\begin{center}%
\begin{tabular}{|l|l|}
\hline
\rbh{Name}\npad{} & \rbh{Description}\dpad{} \\
\hline
\raisebox{0mm}[3mm]{}%
input  & Input $3000$ image PCA coefficients \\
fc1    & Fully connected $3000\to2000$, ReLU activation \\
drop1  & Dropout, $p=0.2$ \\
fc2    & Fully connected $2000\to1000$, $\tanh$ activation \\ 
drop2  & Dropout, $p=0.2$ \\
fc3    & Fully connected $1000\to160$, linear activation \\
output & Fully connected $160\to\nout{}$, linear activation \\
\hline
\end{tabular}%
\end{center}%
\vpad%

The position and the orientation of the head in the input images varies, which in practice necessitates
stabilizing the input images prior to taking the PCA transformation. For this
we used the facial landmark detector of Kazemi and Sullivan~\shortcite{kazemi2014}.
Rotation angle and median line of the face were estimated from the landmark points surrounding
the eyes. Because these were found to shift vertically during blinking of the eyes, the vertical position of the face was determined from
the landmark points on the nose. The image was then rotated to a fixed orientation, and translated so that the point midway between
the eyes remained at a fixed position.

Even though the network may seem overly simplistic, similarly to the all-convolutional architecture, we did not find a way to improve the results by adding more layers or changing
the widths of the existing layers. We experimented with different regularization schemes, but simply adding two dropout layers was
found to yield the best results. The output layer is initialized using a PCA basis for the output meshes computed as in the convolutional
network.

Ultimately, the only aspect in which the fully connected network remained superior to the convolutional network was training time.
Whereas the convolutional network takes~8--10 hours to train in a typical case, the fully connected network
would converge in as little as one hour. Even though it was initially speculated that fast training
times could be beneficial in production use, it ended up not mattering much as long as training could be
completed overnight.

One disadvantage of using a fully connected network is that stabilizing the input images for the fully connected network turned out to be problematic because of residual
motion that remained due to inaccuracies in the facial landmark detection.
This residual jitter of input images sometimes caused spurious and highly distracting motion of output vertices. We
tried hardening the fully connected network to this effect by applying a similar jitter to inputs during training in order to present the
same stabilized input image to the network in slightly different positions, but this did not help.

We suspect that it may be too difficult for the fully connected network to understand that slightly
offset images should produce the same result, perhaps partially due to the input PCA transform. Nonetheless,
the results with input PCA transform were better than using the raw image as the input.

On the other hand, the convolutional network, when trained with proper input augmentation (Section~\ref{sec:augment}),
is not sensitive to the position and orientation of the actor's head in the input images. Hence the convolutional network carries an advantage in that  no
image stabilization is required as a pre-process.
\figconvergence
\figconvergencefc

We see in Figures ~\ref{fig:convergence} and  ~\ref{fig:convergencefc} that the fully connected network often produced numerically better results than the convolutional network,
but visually the results were significantly worse as the fully connected network appeared to generally attenuate the facial motion. 
Even in individual shots where the fully connected
network produced a numerically clearly superior result, the facial motion was judged to lack expressiveness
and was not as temporally stable compared to the results produced by the convolutional network.
We further discuss this general discrepancy between numerical and visual quality below.

\subsection{Training Process Evaluation}

In Section ~\ref{sec:augment} we described several data augmentations we performed that made the network more resistant to variations in input data and eliminated the need for stabilization as a pre-process for our all-convolutional network.
Additionally, we also tried augmenting the data by adding noise to the images and 
applying a variable gamma correction factor to approximate varying skin glossiness.
However, both of these augmentations were found to be detrimental to learning.
Similarly, small 2D perspective transformations---an attempt to crudely mimic the non-linear effects of
head rotations---were not found to be beneficial.

\subsection{Numerical Results}

\figrmseplot
\figvalshotb

Figure~\ref{fig:convergence} shows the convergence of the network for Character~1, trained using 15173 input frames. 
The training set for Character~2 contained 10078 frames.
As previously explained, our loss function is the MSE between the network output
and target positions from the training/validation set. The vertex coordinates are measured in centimeters in our data,
so the final validation loss of 0.0028 corresponds to RMSE of 0.92 millimeters. 
With longer training the training loss could be pushed arbitrarily close to zero, but this did not
improve the validation loss or the subjective quality of the results.

Figure~\ref{fig:valshotb}
 illustrates the numerical accuracy of our trained network on
a selection of interesting frames in validation shots that were not used in training. Note that the RMSE of the frames shown in the figure are higher than average since the validation data mostly consist of more neutral material than the frames shown in the figure. Per-frame RMSE plot
for the validation shots for Character~1 is shown in Figure~\ref{fig:rmseplot}.

We found that the neural network was very efficient in producing consistent output even when 
there were variations in the input data because of inaccuracies in the conventional capture pipeline.
In the first four rows of Figure~\ref{fig:valshotb}, we can see that,
especially in the regions around the hairline and above the eyebrows, 
the target data obtained
from the conventional capture pipeline 
sometimes contained systematic errors 
that the capture artists did not
notice and thus did not correct. 
Because a neural network only learns the consistent
features of the input-output mapping 
as long as overfitting is avoided, 
our network output does not fluctuate in the same way as the manual target positions do.
In fact, visually it is often not clear whether the manually tracked target positions or the inferred positions are closer to the ground truth.
We believe this
explains some of the numerical discrepancies between our output and the validation data.

Given the inevitable variability in the manual work involved in using the conventional capture pipeline, 
we could not hope that our network would reach a numerically exact match with manually prepared validation
data. The goal of performance capture is to generate believable facial motion,
and therefore the perceptual quality of the results---as judged by professional artists---is ultimately
what determines whether a capture system is useful or not in our production environment.

\subsection{Comparison}

\figcomparison

We visually compare our method in Figure~\ref{fig:compare} to Thies et al. \shortcite{Thies2016} and Cao et al.~\shortcite{Cao2014}, two state-of-the-art monocular real-time facial performance capture methods that do not require a rig. 
Since the comparison methods generalize to any identity and our method assumes the identity of one user, in order to make the comparison more fair, we use the method of Li~et~al.~\shortcite{li2010example} to fix the identity mesh for the comparison methods and only track the expressions.
Visually our method appears to be more accurate than \cite{Thies2016} and \cite{Cao2014}, but we note that they bear significant advantages in that they do not require per-user calibration and allow for less constrained head movements.
We also note that if we allowed the comparison methods to retrain for new identities and restricted head movement in all their inputs, their accuracy could be improved to more closely match our levels.
An advantage our method poses over the comparison methods is that it is capable of inferring plausible animations for self-occluded or difficult to track regions such as details surrounding the mouth and eyes.
In a production setting where we have resources to constrain the head movement and perform per-user training and would like to capture the user as accurately and plausibly as possible across all regions of the head, our method is advantageous over other existing methods.
\subsection{Performance}

\begin{table}[t]
\begin{center}
\begin{tabular}{|c|c|c|c|c|}
\hline
Method& Cao et & Thies et & Our method & Our method \\
&  al.~\shortcite{Cao2014} &al.~\shortcite{Thies2016} & (online) & (batched)\\
\hline
Frames/s&28& 28 &287 & 870\\
\hline
\end{tabular}
\vspace*{2mm}
\caption{Throughput comparison with other facial performance capture methods.} \label{runtimes}
\end{center}
\end{table}

Our system runs comfortably in real-time.
As seen in Table ~\ref{runtimes},
we achieve 287 frames per second when used online and up to 870 frames per second if batch processing is used offline with a batch size of 200 frames.  Meanwhile, other real-time methods are only able to achieve 28 frames per second. This allows our system to process a large amount of footage in a short amount of time.

\subsection{Limitations}

We have proposed a system that can achieve high levels of accuracy for facial performance capture while drastically reducing the amount of manual work involved in a production setting. However, we have observed several limitations in our system and suggest future work we can explore.

\paragraph{Non-Optimal Loss Function.} In validation shots for which numerical quality results could be computed, 
the visual quality of the network output did not always follow the value of the loss
function. For example, a shot with a low loss value might exhibit unnatural
movement of lips or eyes, whereas a shot with a higher loss value may look more believable.
This suggests that our current loss function does not get the optimal results possible from a deep neural network, and it should be beneficial to design a more perceptually oriented
loss function for facial expressions similar in spirit of how structural
similarity metrics have been developed for comparing images \cite{Wang2004}.
It seems clear that a better loss function would result 
in a more capable network, as it would learn to focus more on areas that require the
highest precision.

\paragraph{Per-User Calibration.} Despite our ability to capture details more accurately than other methods, one strong limitation of our method compared to other state-of-the-art monocular facial performance capture methods is that we require performing a per-user calibration of retraining the network for each new identity. In the future, we would like to further reduce the amount of manual labor involved in our pipeline and create a system that can achieve the same level of accuracy while also generalizing to all users.

\section{Conclusion}
\label{sec:conclusion}

We have presented a neural network based facial capture method that has proven
accurate enough to be applied in an upcoming game production based on thorough
pre-production testing while also requiring much less labor than other current facial performance capture pipelines used in game production.
Another advantage our method holds in the production setting is that building the dataset for the network enables
tracking work to start any time, so pick-up shoots and screenplay changes are much faster to deliver with high quality.

We have evaluated our network architecture and training pipeline against other network and pipeline variations, and we determined the proposed architecture and augmentation methods to yield a very good balance between optimal visual results and reasonable training time for production purposes.

We have also shown our method to surpass other state-of-the-art monocular real-time facial performance capture methods in our ability to infer a plausible mesh around regions that are invisible or difficult to track such as the area surrounding the eye and mouth. However, our system has a significant drawback as we require per-user calibration. 
The 5--10 minute dataset required for each new identity for high-quality output typically means that the actor needs an important enough
role in the game to justify the cost.

Even though the convolutional network may seem like an opaque building block, our
approach retains all of the artistic freedom because we output simple 3D point
clouds that can be further edited using standard tools, and compressed into standard
character rigs. We feel that many other aspects of the production pipelines of
modern games could benefit from similar, selective use of deep learning for bypassing
or accelerating manual processing steps that have known but tedious or expensive solutions.

\paragraph{Future Work.} Future work may include addressing the limitations of our system mentioned earlier and developing a more accurate pipeline that does not require per-user calibration.
Additional work will also focus on capturing datasets using helmet-mounted cameras for true performance capture of the
face and body simultaneously. Nevertheless, we have presented a system that has drastically reduced the amount of manual work in high quality facial performance capture, and our system represents an important step in the direction of fully automated, high quality facial and body capture.

\bibliographystyle{ACM-Reference-Format}
\bibliography{paper}

\end{document}